\documentclass[journal,transmag]{IEEEtran}
\usepackage{times}
\usepackage{epsfig}
\usepackage{graphicx}
\usepackage{amsmath}
\usepackage{amssymb}
\usepackage{float}
\ifCLASSINFOpdf
\else
\fi
\begin{document}

\title{LGLG-WPCA: An Effective Texture-based Method for Face Recognition}

\author{\IEEEauthorblockN{
Chaorong Li\IEEEauthorrefmark{1,2},
Wei Huang\IEEEauthorrefmark{1},and
Huafu Chen\IEEEauthorrefmark{1}}
\IEEEauthorblockA{\IEEEauthorrefmark{1}University of Electronic Science and Technology of China, Chengdu 611731, China}
\IEEEauthorblockA{\IEEEauthorrefmark{2}Department of Computer  Science  and  Information Engineering, Yibin University, Yibin 644000}
\thanks{This work is supported by the China Postdoctoral Science Foundation (No.2016M602675), Foundation of the Central Universities in China(No.ZYGX2016J123), Project of Education Department of Sichuan Province (No.16ZA0328), Project of Sichuan Science and Technology Program (No.2018JY0117), Science and Technology Project of Yibin of China (2016).(email: lichaorong88@163.com).}}


\IEEEtitleabstractindextext{%
\begin{abstract}
In this paper, we proposed an effective face feature extraction method by Learning Gabor Log-Euclidean Gaussian with Whitening Principal Component Analysis (WPCA), called LGLG-WPCA. The proposed method learns face features from the embedded multivariate Gaussian in Gabor wavelet domain; it has the robust performance to adverse conditions such as varying poses, skin aging and uneven illumination. Because the space of Gaussian is a Riemannian manifold and it is difficult to incorporate learning mechanism in the model. To address this issue, we use L$^2$EMG\cite{Li2017Local} to map the multidimensional Gaussian model to the linear space, and then use WPCA to learn face features. We also implemented the key-point-based version of LGLG-WPCA, called LGLG(KP)-WPCA. Experiments show the proposed methods are effective and promising for face texture feature extraction and the combination of the feature of the proposed methods and the features of Deep Convolutional Network (DCNN) achieved the best recognition accuracies on FERET database compared to the state-of-the-art methods. In the next version of this paper, we will test the performance of the proposed methods on the large-varying pose databases.
\end{abstract}

}

\maketitle

erence papers position the abstract like regular
\IEEEdisplaynontitleabstractindextext
\IEEEpeerreviewmaketitle

\section{Introduction}

Because of the potential application value, face recognition has always been a research hotspot in the field of machine vision, such as identity authentication, access control system and online transaction. However, unfavorable factors constitute great challenges to the automatic face recognition. These adverse factors include: image noise, low resolution image, uneven illuminations and varying poses, skin aging, facial expressions and facial occlusion. Among these factors, uneven illuminations, facial occlusion and varying poses are most prone to occur in unconstrained environments.

To face recognition, the effects of uneven illumination can be counteracted by conventional image processing techniques like the simple and efficient preprocessing chain \cite{Tan2010Enhanced}. However, facial occlusion and varying pose are more difficult than uneven illuminations, because the conventional image processing methods cannot correct or reconstruct a high fidelity face image for the damaged image\cite{cfppaper}.

\begin{figure}\label{preprocessing}
  \centering
  \includegraphics[width=3.5in]{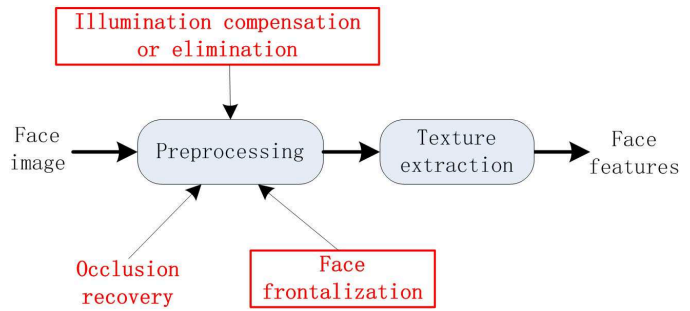}
  \caption{Preprocessing texture feature extraction method}
\end{figure}

In general, face recognition methods against adverse factors fall into two categories. (1) Calculate the face features directly from face image using Deep Convolutional Network (DCNN) based methods such as DeepID\cite{Sun2014Deep}, FaceNet\cite{Schroff2015FaceNet}, VGGFace\cite{Parkhi2015Deep} and PCANet\cite{Chan2015PCANet}. These methods use a large number of samples to adapt to the face recognition in different poses, different expressions, and lighting. (2) Preprocessing-based feature extraction method (see Figure\ref{preprocessing}), which is the s scheme adopted by most face recognition methods. In preprocessing-based method, the preprocessing step is firstly used to remove or counteract the advertise factors such as noise and varying pose, and then texture feature extraction descriptors (like LBP\cite{ahonen2006face}, LTP\cite{Tan2010Enhanced}, SIFT\cite{Geng2010Face} HOG\cite{D2011Face} and Gabor\cite{yu2010gabor}) or DCNNs are used to produce the face features.

Recently, a few effective preprocessing-based methods have been developed to deal with the most common and difficult large-varying-pose problem: For example, DeepFace\cite{Taigman2014DeepFace} uses 3D to align faces; TP-GAN\cite{Huang2017Beyond} uses Generative Adversarial Nets(GAN) \cite{Goodfellow2014Generative} to rectify (frontalize) face image and Light CNN\cite{He2017Learning} for face recognition; HF-PIM\cite{Cao2018Learning} combines GAN and 3D Morphable Model\cite{Blanz1999A} to frontalize the face and then use Light CNN to extract facial features; DR-GAN\cite{Luan2017Disentangled} trains the Encoder-Decoder network while using GAN to frontalize the faces. Large varying poses or occlusions may be corrected (frontalized) by using 3D Morphable model or GAN. However, the 3D Morphable or GAN based methods cannot rectified the small varying expressions or varying poses and moreover, the rectified images will produce more or less deviation compared with the ground truth images.

In fact, because small varying expressions and poses are inevitable in the applications of face recognition, it is important that feature extraction method needs to have robust performance to the small variations. In this paper, we propose a preprocessing-based texture feature extraction method (called LGLG-WPCA) for face recognition. LGLG-WPCA extracts face features using multidimensional Gaussian models on Gabor wavelet domain; it is a nice texture feature descriptor and is insensitive to noise and is also robust to small varying poses and expressions. Previous methods \cite{pang2008gabor,li2019marginal} construct the covariance matrixes to represent image and these methods ignore the mean information of the subbands of Gabor wavelet. Multidimensional Gaussian model is an extension of the covariance matrix, and it contains more information than covariance matrix. However, both covariance matrix and multidimensional Gaussian model also belong to Riemannian manifold. When comparing two covariance-matrix based models in Riemannian space, the computational cost is much higher than that of comparing two vectors in Euclidean space; furthermore, it is difficult to incorporate a learning mechanism to improve the performance of the models in Riemannian space. In order to reduce the computational cost and improve the performance, we use L$^2$EMG\cite{Li2017Local} to embed multivariate Gaussian model into Euclidean space and use Whitening Principal Component Analysis (WPCA) to learn discriminative face features from the embedded Gaussian model.
\section{Related work}
Gabor wavelet (Gabor filters) is a useful texture extraction tool in the field of computer vision. in a number of researches. Face recognition methods based on Gabor wavelet have been reported in large numbers\cite{Liu2002Gabor, Zhang2005Local, Shen2007Gabor, Zhenhua2014Gabor}. Because Gabor wavelet is a very redundant transform, researchers used Principal Component Analysis (PCA) or Liner Discriminant Analysis (LDA) to compress the Gabor subbands. For example, Gabor-Fisher classifier\cite{Liu2002Gabor, Shen2007Gabor} applied the Fisher linear discriminant to obtain the discriminative face feature from Gabor wavelet subbands which are yielded from image face; Gabor-wavelet based kernel PCA\cite{Liu2004Gabor} used kernel PCA to compress the Gabor wavelet subbands. Some researchers\cite{Zhang2005Local, Zhenhua2014Gabor} encoded the Gabor subbands by using Local Binary Pattern (LBP), and then obtained the face features.

Means and standard-deviations of the Gabor wavelet subbands (Gabor features) are the discriminative information of images, and they can be used as the image features; however, the low-dimensional features composed of means and standard-deviations are difficult to distinguish between different faces which are highly similar between-classes and large varying within-classes. Compared with means and standard-deviations, covariance matrix is an effective measure for augmenting standard deviation features. Given the number of Gabor wavelet subbands is $d$, the dimensionality of mean and standard-deviation features is $2d$, whereas the size of covariance matrix features is $d\times d$ and the different value of covariance matrix is $(d^2+d)/2$. Therefore, covariance matrix contains more discriminative information than that of the features composed of mean and standard deviation.

Covariance matrix has been widely used to image representation\cite{tuzel2006region, pang2008gabor, Wang2012Covariance, Minh2016Kernel, Wang2017CVPR, Zhang_2018_CVPR}. In \cite{tuzel2006region}, Tuzel et al. map each pixel of image to a 5-dimensional feature space (intensities, norm of first and second order derivatives of intensities), and use a covariance matrix to model these features. Yanwei et al.\cite{pang2008gabor} proposed a covariance-matrix based method to model Gabor subbands; in their work, both pixel locations and Gabor features are employed to construct the covariance matrices. Wang et al.\cite{Wang2012Covariance} use covariance matrix for image set based face recognition. Recent years, covariance or covariance matrix are introduced into deep learning-based networks\cite{Covariance-Pooling, Li_2018_CVPR}. Because covariance matrix is Riemannian manifold, Euclidean distance cannot be directly used as the measure of covariance matrix. Therefore, Tuzel et al.\cite{tuzel2006region} proposed Riemannian distance as the measure of covariance matrix. Given two covariance matrices $\pmb C_1$ and $\pmb C_2$, Riemannian distance is defined as:
\begin{equation}\label{RD}
   RD\left (\pmb C_1, \pmb C_2 \right )=\sqrt{\sum_{i=1}^{d}ln\lambda_i^2\left (\pmb C_1, \pmb C_2 \right )},
\end{equation}
where $\lambda_i \left (\pmb C_1, \pmb C_2 \right )_{i=1,\cdots,d}$ are the generalized eigenvalues of $\pmb C_1$ and $\pmb C_2$. The computation cost of Riemannian distance in the feature matching step is expensive because we should calculate the generalized eigenvalues of $ \pmb C_1$ and $\pmb C_2$ in (\ref{RD}). Researchers have developed a embedding approach which transforms the covariance matrix into a linear space, called Log-Euclidean distance (LED)\cite{Arsigny2011Geometric, Arsigny2006Log}. LED is expressed as:
\begin{equation}\label{LDx}
   LED\left (\pmb C_1, \pmb C_2 \right )={\left \|log(\pmb C_1)-log(\pmb C_2)\right\|}_F,
\end{equation}
where log is matrix logarithm operator and ${\left \| \cdot \right \|}_F$ denotes the Matrix Frobenius Norm (MFN). Besides, Minh et al.\cite{Minh2016Kernel} provided a finite-dimensional approximation of the Log-Hilbert-Schmidt (Log-HS) distance between covariance operators to image classification.

Covariance matrix is a special case of multivariate Gaussian distribution which parameters consist of mean vector and covariance matrix. Similar to covariance matrix, the space of Gaussian is not a linear space but a Riemannian manifold. Peihua et al. \cite{Li2017Local} used Lie group to embed multivariate Gaussian in to Euclidean space, called Local Log-Euclidean Multivariate Gaussian (L$^2$EMG). Embedding form of Gaussian Euclidean space is denoted by
\begin{equation}\label{L2EMG1}
 \pmb B=log\begin{bmatrix}
\pmb C+\pmb\mu\pmb\mu^T & \pmb\mu\\
\pmb\mu^T & 1
\end{bmatrix}^{\frac{1}{2}}
\end{equation}
where $\pmb C$ and $\pmb \mu$ are the covariance matrix and mean parameters of Gaussian. The similarity of two Gaussian models is denoted by
\begin{equation}\label{L2EMG2}
  L^2EMG(\pmb C_1,\pmb C_2)=\left \| \pmb B_1-\pmb B_2 \right \|_F
\end{equation}


Multivariate statistical models including covariance-based methods have two shortcomings. First, because of the complicated matrix operation, the computation of the measure in a Riemannian space is time-consuming compared to the computation in a Euclidean space. Second, it is difficult to resort to an effective learning approach to improve the performance. Recently a few methods have been proposed to address these shortcomings. For example, Harandi et al. \cite{Harandi2017Dimensionality} use an orthonormal projection model to project the high-dimensional manifold to a low-dimensional vector. Wang et al.\cite{Wang2017CVPR} developed a discriminative covariance-oriented representation learning framework to deal with face recognition. Different from the methods mentioned above, we proposed an efficient method by WPCA to learn the face features from the multivariate Gaussian in Gabor domain. Our methods can efficiently describe the texture feature of face and it is robust to small variance of face such as facial expressions.

\section{Gabor wavelet}
\begin{figure*}[!htbp]
\centering
  \includegraphics[width=6in]{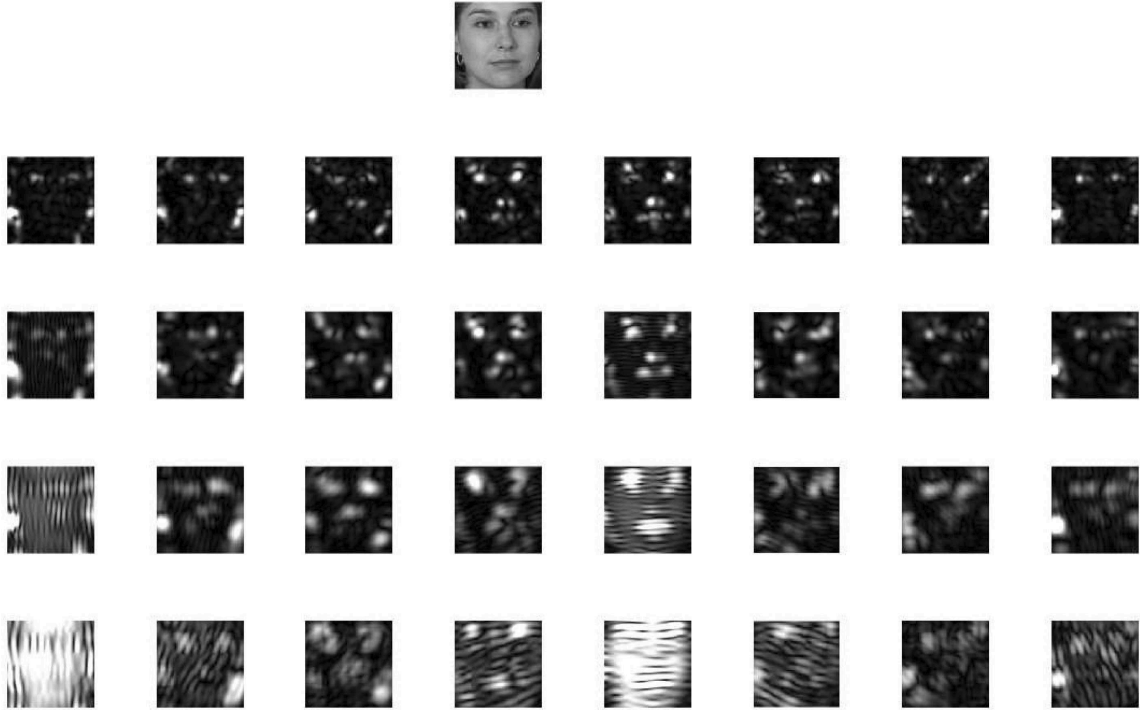}\\
  \caption{Gabor wavelet decomposition.}\label{gaborFace}
\end{figure*}
Gabor Wavelet is an efficient tool for image analysis. It is defined as the convolution on the image with a set of Gabor filters. The 2D Gabor filter is the product of a Gaussian function and the complex exponential function, denoted by\cite{Liu2002Gabor, pang2008gabor}
\begin{equation}\label{gaborEQ2}
 \psi_{u,v}\left ( z \right ) =\frac{\left \| k_{u,v} \right \|^2}{\sigma^2}e^{-\frac{\left \| k_{u,v} \right \|^2\left \| z \right \|^2}{2\sigma^2}}\left [ e^{ik_{u,v}z}- e^{-\frac{\sigma^2}{2}} \right]
\end{equation}
where $z=(x,y)$; $u$ and $v$ define the direction and scale of the Gabor filters (Therefor, $\psi_{u,v}$ is called Gabor wavelet), $ \left \| \cdot  \right \|$ denotes the norm operator; $k_{u,v}$ is the wave vector which has the following express
\begin{equation}\label{kuv}
  k_{u,v}=k_v e^{i \psi_u}
\end{equation}
where $k_v =k_{max}/f^v$, $k_{max}$ is the maximum frequency; $\psi_v=\frac{\pi u}{8}$; $f$ is the spacing factor between kernels in the frequency domain. In this work, $f=\sqrt 2$ and $u$ takes the values from set $\left\{1, 2,\cdots, U \right\}$ and $v$ takes the values from set $\left\{1, 2,\cdots, V \right\}$, where $U$ and $V$ are the maximum number of directions and scales. Given $U$-direction and $V$-scale of Gabor wavelet is performed on an image, there will be producted $U \times V$ responses (subbands). The subbands of the Gabor wavelet are complex and the amplitudes of subbands are used in this paper. Fig.\ref{gaborFace} shows an 8-direction and 4-scale decomposition of Gabor wavelet.
\section{Learning Gabor Log-Euclidean Gaussian (LGLG) for face recognition}
In the proposed method, Gabor wavelet is used to extract the texture feature of face. However, there are very a few number subbands produced by Gabor wavelet. Previous work use covariance matrix to capture the subbands\cite{pang2008gabor}. However, covariance matrix ignores the means information of subbands. LGLG uses multivariate Gaussian which contains both the covariance and means to model the subbands. There are two steps in LGLG. The first step is to extract the local feature using Gabor Log-Euclidean Gaussian (GLG); the second step is to Learn GLG (LGLG) for face recognition.
\subsection{Extract local feature using Gabor Log-Euclidean Gaussian}
\begin{figure*}[!htbp]
  \centering
  \includegraphics[width=7in]{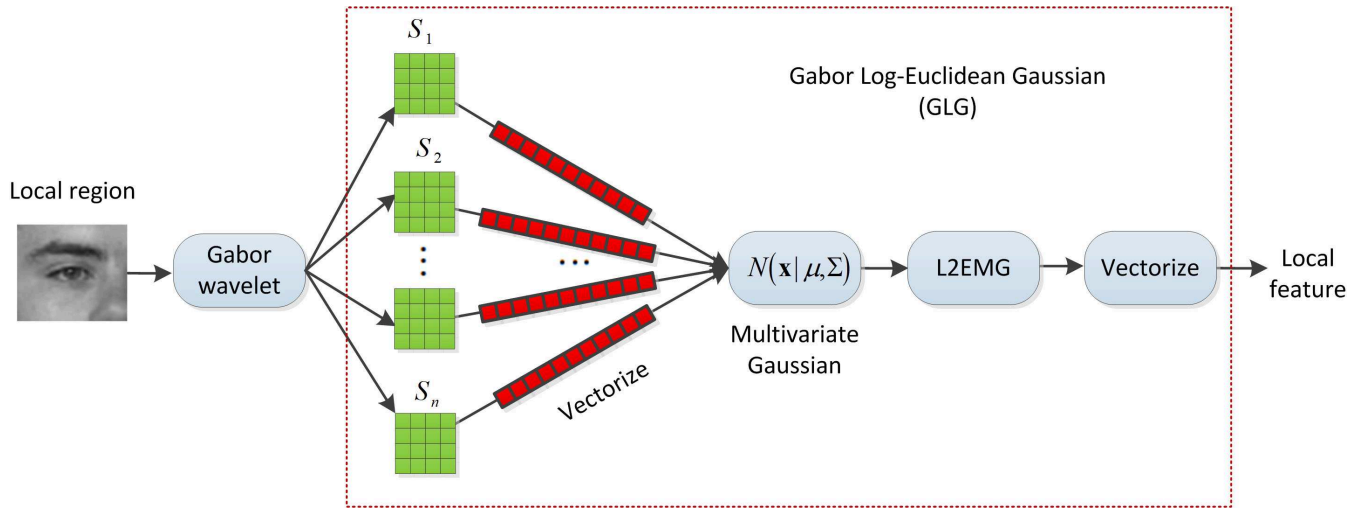}\\
  \caption{Extract local feature using GLG.}\label{GLGFig}
\end{figure*}
The detailed scheme of constructing a GLG on a block of face is shown in Fig.\ref{GLGFig}. Gabor wavelet is first used decompose the local block of face image and each of the decomposed magnitude subband of Gabor wavelet is vectorized into 1-dimension vector, denoted by $\pmb x =\left \{ x_1, x_2, \cdots, x_L \right \}$, where $L$ is the length of $\pmb x$. To perform a  $U$-direction and $V$-scale decomposition, $P$ ($P=U \times V$) 1-dimension vectors are to be yielded.  Then all of vectors produced from the subbands are formatted as following matrix:
\begin{equation}\label{vectorX}
\pmb X=\left[ \pmb x_1, \pmb x_2, \cdots, \pmb x_P \right]
\end{equation}
If $ \pmb X$ is regarded as observation matrix of a random vector and each column of the matrix corresponds to the observations of a random variable, then $X$ is approximately Gaussian distribution \cite{Li2017Color}. We can calculate the parameter covariance matrix $\pmb C$ and mean $\pmb m$ of Gaussian on $ \pmb X $ by using maximum Likelihood Estimation (MLE).
\begin{equation}\label{muQ}
  \pmb \mu=\frac{1}{P}\sum_{k=1}^{P} \pmb x_k
\end{equation}
\begin{equation}\label{covQ}
  \pmb C=\frac{1}{P}\sum_{k=1}^{P} \left (\pmb x_k-\mu \right) \left (\pmb x_k-\mu \right)^T
\end{equation}
According to the estimated $\pmb C$ and $\mu$, we use EQ.\ref{L2EMG2} to embed the Gaussian model constructed from subbands of Gabor wavelet in Euclidean space and vectorize it into a local feature vector, denoted by
\begin{equation}\label{LogGaborGaussian}
  F_i=vec \left ( log\begin{bmatrix}
\pmb C+\pmb\mu\pmb\mu^T & \pmb\mu\\
\pmb\mu^T & 1
\end{bmatrix}^{\frac{1}{2}} \right )
\end{equation}
where $i={1, 2, \cdots, P}$. Sign $vect (\cdot)$ denotes the vectorization of matrix.
\subsection{Learning GLG using WPCA (LGLG-WPCA) for face recognition}
\begin{figure*}[htbp]
  \centering
  \includegraphics[width=6in]{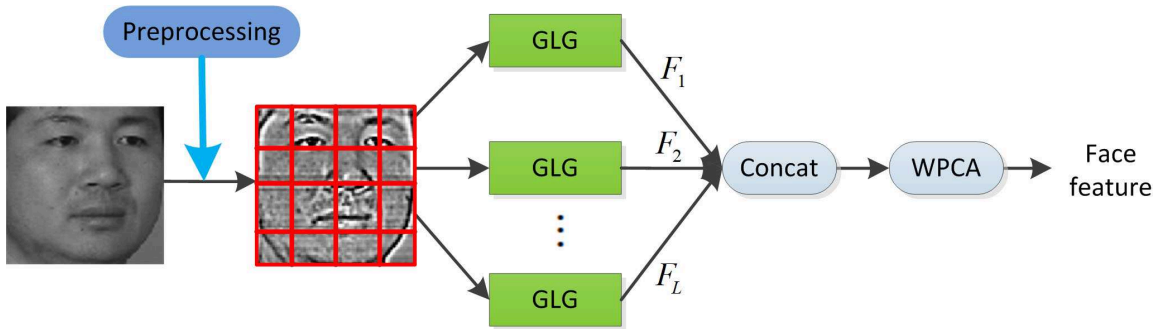}\\
  \caption{Learning Gabor Log-Euclidean Gaussian using WPCA (LGLG-WPCA) for face recognition} \label{LGLG}
\end{figure*}

We learning GLG using WPCA for face recognition, called LGLG-WPCA. In LGLG-WPCA, we use three preprocessing approaches (Gamma correction, Difference of Gaussian filter and contrast normalization) described in \cite{Tan2010Enhanced} to counter the effects of illumination variations. Before extracting face features, the image is divided into $m\times n$ local square blocks and GLG is applied to all the blocks producing $L \left\{ L=m\times n \right\}$ vectors denoted by $F_1, F_2, \cdots, F_L $. The feature vectors $\pmb F_{i}$ of the local square blocks in the image are concatenated into a high-dimensional vector $\pmb F$, and Whitening Principal component analysis (WPCA) is used to project the high-dimensional vector into a low-dimensional feature vector. The flowchart of LGLG-WPCA is shown in Fig.\ref{LGLG}

WPCA is more efficient than Principal component analysis (PCA) for face recognition under the condition of the training set has single sample per person. Compared to PCA, the extra benefit of WPCA is it normalizes the contribution of each principal component by using whitening transformation which divides the principal components by standard deviations.

The columns of $\pmb U$ are composed of the eigenvectors of the covariance matrix. A high-dimensional vector can be compressed into a low-dimensional vector $\pmb y$ by projecting it on $\pmb U$, that is:
\begin{equation}\label{pcaf1}
  \pmb y=\pmb U^T \pmb x.
\end{equation}
The second step of WPCA is to transform the $\pmb U$ into $\pmb W$ by using whitening transformation:

\begin{equation}\label{WPCAF}
   \pmb W=\pmb U \left(\pmb D\right)^{-\frac{1}{2}},
\end{equation}
where $\pmb D=diag \{\lambda_1, \lambda_2, \cdots\}$. Then the projected WPCA features $\pmb y$ are
 \begin{equation}\label{pcaf1}
  \pmb y=\pmb W^T \pmb x=\left(\pmb U \left(\pmb D\right)^{-\frac{1}{2}}\right)^T \pmb x.
\end{equation}

However, WPCA may suffer performance degradation problem when the eigenvalues of the covariance matrices are very small or close to zero. If the eigenvalues of covariance matrix are too small and we use WPCA to whitening the features, it will over-amplify the influence of the small eigenvalues in feature matching. To address the issue, we standardize the WPCA features with z-score standardization (ZSCORE), which is denoted as:
\begin{equation}\label{equationZ}
  \pmb z=\frac{\pmb y-MEAN(\pmb y)}{STD(\pmb y)},
\end{equation}
where MEAN$(\pmb y)$ and STD$(\pmb y)$ denote the mean and standard deviation of $\pmb y$, respectively. In face recognition, standard Euclidean distance is used as the similarity between two face images.
\begin{figure*}[htbp]
  \centering
  \includegraphics[width=6in]{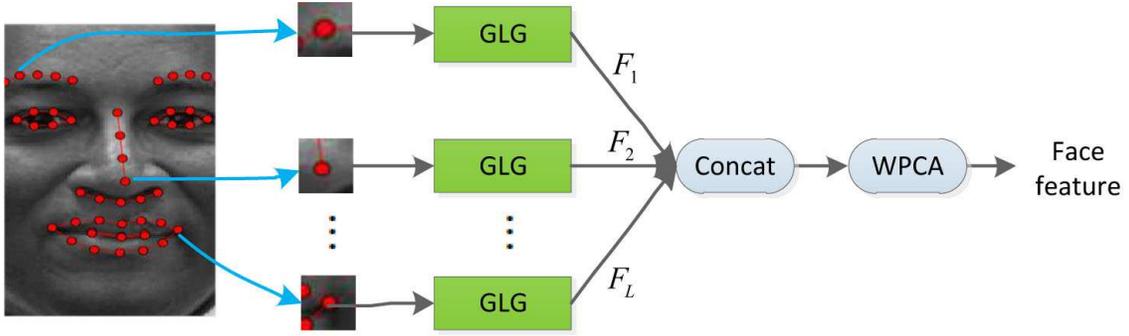}\\
  \caption{Learning Gabor Log-Euclidean Gaussian using WPCA based on key points (LGLG(KP)-WPCA) for face recognition.}\label{LGLGKP}
\end{figure*}

In addition, we can also use LGLG to extract the face features based on local regions centered on corresponding key points. LGLG-WPCA based on key points is called LGLG(KP)-WPCA. Fig.\ref{LGLGKP} shows the flow chart of LGLG(KP)-WPCA. In LGLG(KP)-WPCA, we  use SMD \cite{xiong2013supervised}, which can yield 49 key points on a face images, to detect the key points and 21 key points are selected. In each key point, we extract a local square block around a key point and in each square block we use LGLG to extract the local feature $F_i$. Similar to LGLG-WPCA, all the local features $F_i$(where $i=1 \cdots L$) are concatenated into a high-dimensional feature vector and WPCA is used to compress the high-dimensional feature vector into a discriminative and low-dimensional feature vector.
\section{Experiments}
\begin{figure}[htbp]
  \centering
  \includegraphics[width=3.2in]{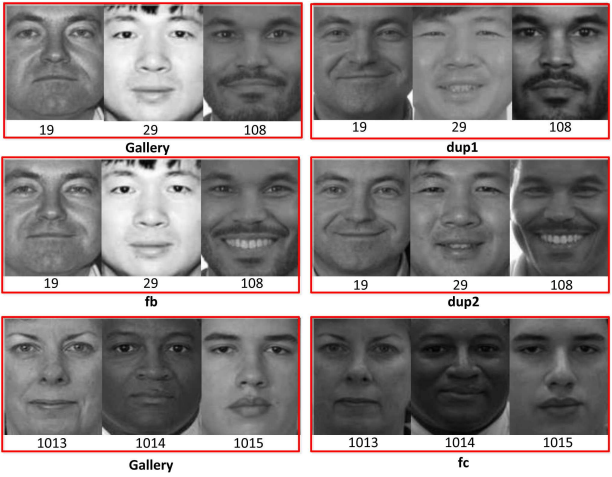}\\
  \caption{Samples selected from gallery subset and the three subsets (fb, fc, dup1 and dup2). The numbers under the corresponding images are the labels of the corresponding persons.}\label{FERETsamples}
\end{figure}
Our evaluation is carried out on standard FERET database. Standard FERET contains four subsets: a gallery, a facial expression subset (fb), an illumination subset (fc), and two duplicate subsets (dup1 and dup2). Dup1 probe images were obtained at different times. The harder dup2 probe subset is a subset of dup1; the images are in dup2 are taken only at least 18 months. Some images selected from these subsets are shown in Fig.\ref{FERETsamples}. Because there are 1196 person in the gallery, we use WPCA to compress the GLG features of all the local square blocks into an 1196-dimension feature vector and Euclidean norm 2 is used as the similarity of two feature vectors.


The parameter of LGLG will affect the performance of recognition. These parameters include the Block Size (B-Size) of the divided local square blocks of images, and the parameters of Gabor wavelet which consist of the $\sigma$, the local Window Length (WinLen), the number of directions and scales of Gabor wavelet. We tested the recognition accuracies of LGLG-WPCA and LGLG(KP)-WPCA  with respect to different parameter combinations via experiments on subset dup2. The test results of LGLG-WPCA and LGLG(KP)-WPCA are listed in Table \ref{tab:parameters} and Table \ref{tab:Keyparameters}.  To LGLG-WPCA, there are two parameter combinations which obtain the best recognition accuracy 97.44\%: WinLen(9) Sig(1.2) Direct(8) Scale(5) B-Size (13$\times$13); WinLen(9) Sig(1.0) Direct(8) Scale(4) B-Size (15$\times$15). In the following experiments we use the second parameter combination for LGLG-WPCA. To LGLG(KP)-WPCA, the best parameter combinations which obtains the best recognition accuracy 91.02\% is: WinLen(9) Sig(1.2) Direct(8) Scale(5) B-Size (22$\times$22).
\begin{table}[!htbp]
  \centering
  \caption{ The recognition accuracy (percent) of LGLG-WPCA with respect to different parameters on subset dup2.}
  \begin{tabular*}{8cm}{@{\extracolsep{\fill}}lccccc}
    \hline
      WinLen & Sig ($\sigma$)& Direct(U) & Scale (V)& B-Size & Acc. \\
       \hline
      11 &	2.0$\pi$	&8	&5	&$13\times13$&95.72\\
      11&	1.8$\pi$&8	&5	&   $13\times13$&95.72\\
      11&   1.4$\pi$	&8	&5	&$13\times13$	&95.72\\
      11&	1.2$\pi$&	8	&5	&$13\times13$	&96.15\\
      11&   1.2$\pi$&	8&	4&	$13\times13$&	95.72\\
      \hline
    9&	2.0$\pi$	&8	&5	&$13\times13$	&95.72\\
    9&	1.4$\pi$	&8	&5	&$13\times13$	&96.15\\
    9&	1.0$\pi$	&8	&5	&$13\times13$	&95.72\\
    \pmb{9}& \pmb{1.2$\pi$}& \pmb{8}& \pmb{5}& \pmb{$13\times13$}&\pmb{97.44}\\

     \hline
    9&	1.8$\pi$	&8	&5	&$11\times11$	&85.89\\
    9&	1.2$\pi$	&8	&5	&$11\times11$	&87.17\\
    9&	1.2$\pi$ & 8	&4	&$11\times11$	&81.19\\
    \hline
     9&	1.8$\pi$	&8	&5	&$15\times15$	&94.87\\
     9&	1.4$\pi$    &8	&5	&$15\times15$	&96.58\\
     9&	1.2$\pi$	&8	&5	&$15\times15$	&97.08\\
     9&	1.0$\pi$&	8	&5	&$15\times15$	&96.15\\

    9&	1.8$\pi$	&8	&4	&$15\times15$	&95.29\\
    9&	1.6$\pi$	&8	&4	&$15\times15$	&95.72\\
    9&	1.2$\pi$	&8	&4	&$15\times15$	&97.08\\

    \pmb{9}&	\pmb{1.0$\pi$}	&\pmb{8}	&\pmb{4}&\pmb{$15\times15$}	&\pmb{97.44}\\
    \hline
    7&	2.0$\pi$&	8	&5	&$13\times13$	&96.15\\
    7&	1.4$\pi$&8	&5	&$13\times13$	&97.44\\
    7&	1.2$\pi$&8	&5	&$13\times13$	&97.08\\
    7&	1.0$\pi$	&8	&5	&$13\times13$	&96.58\\
    7&	0.8$\pi$	&8	&5	&$13\times13$	&95.72\\
 \hline
  \end{tabular*}
  \label{tab:parameters}
\end{table}

\begin{table}[!htbp]
  \centering
  \caption{ The recognition accuracy (percent) of LGLG(KP)-WPCA with respect to different parameters on subset dup2.}
  \begin{tabular*}{8cm}{@{\extracolsep{\fill}}lccccc}
    \hline
      WinLen & Sigma ($\sigma$)& Direct(U) & Scale (V)& B-Size & Acc. \\
       \hline
      11&2.0$\pi$	&8	&5	&$26\times26$	&87.17\\
    11	&1.8$\pi$	&8	&5	&$26\times26$	&87.17\\
    11	&1.2$\pi$	&8	&5	&$26\times26$	&86.32\\
      \hline
    9	&1.2$\pi$	&8	&5	&$26\times26$	&89.74\\
    9	&1.8$\pi$	&8	&5	&$26\times26$	&90.6\\
    9	&2.0$\pi$	&8	&5	&$26\times26$   &89.74\\
     \hline
    9	&1.8$\pi$	&8	&5	&$22\times22$	&90.17\\
    \pmb{9}	&\pmb{1.2$\pi$}	&\pmb{8}	&\pmb{5}	&\pmb{$22\times22$}	&\pmb{91.02}\\
    9	&1.2$\pi$	&8	&4	&$22\times22$   &88.88\\
    \hline
    9	&1.8$\pi$	&8	&5	&$30\times30$	&88.88\\
    9	&1.8$\pi$	&8	&4	&$30\times30$	&87.17\\
    9	&1.6$\pi$	&8	&4	&$30\times30$   &87.6\\
    9	&1.2$\pi$	&8	&4	&$30\times30$	&88.46\\
    9	&1.0$\pi$	&8	&4	&$30\times30$	&86.75\\
    \hline
    7	&2.0$\pi$	&8	&5	&$26\times26$   &87.17\\
    7	&1.5$\pi$	&8	&5	&$26\times26$	&88.46\\
    7	&1.2$\pi$	&8	&5	&$26\times26$	&88.88\\
    7	&1.0$\pi$	&8	&5	&$26\times26$	&88.03\\
    7	&0.8$\pi$	&8	&5	&$26\times26$   &83.33\\
   \hline
  \end{tabular*}
  \label{tab:Keyparameters}
\end{table}

We compared the proposed method LGLG with the state-of-the-art methods including texture descriptors (such as LBP based method MDML-DCPs+WPCA\cite{ding2016multi} and Gabor based method LGBP+LGXP+LDA\cite{Xie2010Fusing}) and DCNN methods (such as VGGFace\cite{Parkhi2015Deep}, PCANet \cite{Chan2015PCANet}, ResNet50\cite{He2016Deep,cao2018vggface2} and SENet\cite{Hu2017Squeeze,cao2018vggface2}), and the recognition results are shown in Table \ref{tab:ssppferet}. The recognition accuracies of LGLG are 99.75\%, 100\%, 97.23\% and 97.44\% on fb, fc, dup1 and dup2, respectively; and its average recognition accuracy reaches 98.60\% which are the best among all the methods. Three Riemannian manifold models in Gabor wavelet domain are implemented for the comparison: COV-GW + RD (Covariance matrix in Gabor Wavelet (GW) domain and Riemannian distance\cite{pang2008gabor} is used), COV-GW + LEG(Covariance matrix in Gabor Wavelet domain with Log-Euclidean embedding and Matrix Frobenius Norm (MFN) is used\cite{Arsigny2011Geometric} as the model's similarity) and GW + L$^2$EMG (We implemented L$^2$EMG \cite{Li2017Local} in Gabor wavelet domain and MFN is used  as the model's similarity). Four DCNN-based methods are used for evaluating our methods in this experiment: PCANet \cite{Chan2015PCANet}, VGGFace\cite{Parkhi2015Deep}, ResNet50\cite{He2016Deep,cao2018vggface2} and SENet\cite{Hu2017Squeeze,cao2018vggface2}.

\begin{figure}[!htbp]
  \centering
  \includegraphics[width=3.2in]{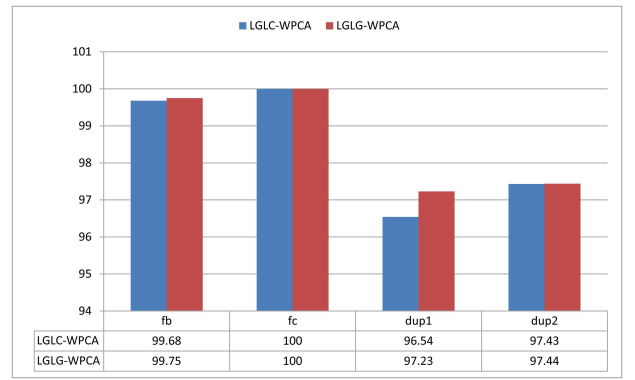}\\
  \caption{The performance comparison between LGLG-WPCA and LGLC-WPCA}\label{LGLGAA}
\end{figure}

It can be observed that LGLG obviously outperforms the three multivariate models improving by about 9 percentage points and has significantly reduction of computational cost compared with Riemannian distance because WPCA is used to reduce the dimensions of features. We also implemented the WPCA learning on the covariance models in Gabor wavelet domain (called LGLC-WPCA). The recognition accuracies of LGLC-WPCA and LGLG-WPCA are shown in Fig.\ref{LGLGAA} and we see that LGLG-WPCA always outperforms the LGLC-WPCA on the three subsets except fb. LGLG(KP)-WPCA is obviously inferior to LGLG-WPCA on the four test subsets. LGLG-WPCA even obtain better accuracies than all the DCNN-based methods. It is well known that DCNN is good at for face recognition under the conditions of large varying illuminations and poses. In order to improve the robust, we combined the LGLG-WPCA/LGLG(KP)-WPCA features and SENet features for face recognition. Because the output of SENet is a 2048-dimensional feature vector, we use WPCA to compress the 2048-dimensional vector into a 480-dimensional vector and concatenate the 480-dimensional vector with the features of LGLG-WPCA (called LGLG-WPCA+SENet-WPCA), as well as the features of LGLG-WPCA(KP) (called LGLG-WPCA(KP)+SENet-WPCA); the number of the final face feature dimensions is 1676 (1196+480). In Table \ref{tab:ssppferet}, the average recognition accuracies of LGLG-WPCA+SENet-WPCA and LGLG-WPCA(KP)+SENet-WPCA are 99.43\% and 99.30\%, respectively. This manifests that the combination of Riemannian features in Gabor wavelet domain and the DCNN features (high level features) can improve the performance of face recognition.
\begin{table*}[!htbp]
  \centering
  \caption{ The recognition accuracy (percent) on standard FERET.}
  \begin{tabular*}{16cm}{@{\extracolsep{\fill}}lccccc}
    \hline
      \quad Method & fb & fc & dup1& dup2 & Avg\\
    \hline
      \quad LBP\cite{ahonen2006face}  & 96.90 & 98.45 & 83.93  & 82.48& 90.44\\
     \quad LTP\cite{Tan2010Enhanced} &96.90 &98.97& 83.93 &83.76 &90.89\\
      \quad LGBP+LGXP+LDA\cite{Xie2010Fusing} & 99.00 &99.00 &94.00 &93.00&96.25\\
     \quad DFD+WPCA \cite{Lei2014Learning} &99.40 &100.0 &91.80& 92.30&95.88\\
      \quad MDML-DCPs+WPCA\cite{ding2016multi} &99.75& 100.0 &96.12 &95.73&97.90\\
     \quad SCBP\cite{Deng2018Compressive}&98.9& 99.0 &85.2& 85.0& 92.03\\
     \quad FFC\cite{Low2018Multi}&99.50 &100 &96.12 &94.87 &97.62\\
    \quad LPOG\cite{Huu2015Local} & 99.8&100&97.4&97.0&98.55\\
      \hline  
     \quad COV-GW + RD\cite{pang2008gabor} & 97.99  & 99.48 & 80.74 &78.21 & 89.11\\
     \quad COV-GW + LEG\cite{Arsigny2011Geometric}&98.07&99.48 & 81.44& 80.34&89.83\\

     \hline
   \quad PCANet \cite{Chan2015PCANet} & 99.58& 100 &95.43 &94.02 &97.26\\
   \quad VGGFace\cite{Parkhi2015Deep} & 98.74&96.39& 86.28&87.61&92.26\\
   \quad ResNet50\cite{He2016Deep,cao2018vggface2} & 99.58&99.49& 96.95&96.58&98.15\\
  \quad  SENet\cite{Hu2017Squeeze,cao2018vggface2}& 99.33& 99.49 &97.22 &97.00 &98.26 \\
   \hline
       \quad \pmb{GW + L$^2$EMG} & \pmb{98.07}& \pmb{99.48} &\pmb{82.13}& \pmb{81.19} &\pmb{90.22}\\
   \quad \textbf{LGLG-WPCA} & \textbf{99.75}&\textbf{100} &\textbf{97.23}&  \textbf{97.44}& \textbf{98.60}\\
   \quad \textbf{LGLG-WPCA(KP)} & \textbf{99.66}&\textbf{99.48} &\textbf{92.38}&  \textbf{91.02}& \textbf{95.64}\\
   \quad \textbf{£¨LGLG-WPCA+SENet-WPCA} & \textbf{99.83}&\textbf{100} &\textbf{98.75}&  \textbf{99.15}& \textbf{99.43}\\
   \quad \textbf{£¨LGLG-WPCA(KP)+SENet-WPCA} & \textbf{99.83}&\textbf{100} &\textbf{98.20}&  \textbf{99.15}& \textbf{99.30}\\
    \hline
  \end{tabular*}
  \label{tab:ssppferet}
\end{table*}
The proposed methods are also computationally efficient (see Table \ref{tab:dimension}). In the experiments, the number of feature dimensions of LGLG-WPCA / LGLG-WPCA(KP) is 1196 which is same as the number of persons in gallery set; the feature dimension of LGLG-WPCA+SENet-WPCA / LGLG-WPCA(KP)+SENet-WPCA is 1676; more importantly, we use Euclidean distance with lower computational cost to recognize faces. Among other Riemannian-manifold-based methods, COV-GW-RD uses Riemannian distance whose computational cost is expensive; COV-GW-LEG and GW-L2EMG use Frobenius norm of matrices with high-dimensional features to recognize faces.
\begin{table*}[htbp]
\centering
  \caption{The feature matching runtime for recognizing all the face images on the two datasets. To the  Riemannian-manifold-based methods we use matrices to describe the features. For example, the COV-GW-RD which dimension $16 \times [40 \times 40]$ indicates the image is divided into 16 blocks and in each block the size of covariance matrix is $[40 \times 40]$.}  
 \begin{tabular*}{13cm}{@{\extracolsep{\fill}}|l|c|c|}
   \hline
       Method &Dimension &Similarity measure\\ [3pt]
   \hline
   COV-GW-RD \cite{pang2008gabor}&$16 \times [40 \times 40]$ &Riemannian distance\\
   COV-GW-LEG \cite{Arsigny2011Geometric}& $25600=16 \times [40 \times 40]$ &Frobenius norm\\
   GW-L2EMG &$26896=16 \times [41 \times 41]$ & Frobenius norm \\
    \hline
   PCANet (WPCA) \cite{Chan2015PCANet} &1000&Cosine distance\\
   VGGFace (WPCA) \cite{Parkhi2015Deep} & 1000& Cosine distance\\

   ResNet50\cite{He2016Deep,cao2018vggface2}&2048&Euclidean distance\\
   SENet\cite{Hu2017Squeeze,cao2018vggface2}&2048&Euclidean distance\\
     \hline

   LGLG-WPCA&1196&Euclidean distance\\
   LGLG-WPCA+SENet-WPCA&1676&Euclidean distance\\
   LGLG-WPCA(KP)+SENet-WPCA&1676&Euclidean distance\\
   \hline
 \end{tabular*}
\label{tab:dimension}
\end{table*}
\section{Conclusions}
We implemented L$^2$EMG in Gabor wavelet domain and WPCA was used to learning the robust features for face recognition. Our method LGLG-WPCA and its variant LGLG(KP)-WPCA, are efficient for extracting the texture features and they also have robust performance under the condition of illumination and small pose and expression variations. LGLG-WPCA and LGLG(KP)-WPCA are superior to the Gabor-based and LBP-based methods and are also computational efficient because Log-Euclidean embedding and WPCA are used for producing the features. It should be pointed out that our methods may not be efficient to the large-varying pose databases such as CFP\cite{sengupta2016frontal} IJB-A\cite{klare2015pushing} and LFW \cite{huang2008labeled} and it cannot compare our LGLG-WPCA with the frontalizing-based methods such as DR-GAN\cite{Luan2017Disentangled} and HF-PIM\cite{Cao2018Learning} on CPF and LFW because the frontalizing preprocessing are not used in LGLG-WPCA. In the future work, we will use the frontalizing approach in the preprocessing step of LGLG-WPCA and combine DCNN features for face recognition under the condition of large-varying poses.

\bibliographystyle{IEEEtran}

\end{document}